\documentclass[journal]{IEEEtran}

\makeatletter
\def\abstract{\normalfont
    \if@twocolumn
      \@IEEEabskeysecsize\bfseries\textit{\abstractname} \relax
    \else
      \bgroup\par\addvspace{0.5\baselineskip}\centering\vspace{-1.78ex}\@IEEEabskeysecsize\textbf{\abstractname}\par\addvspace{0.5\baselineskip}\egroup\quotation\@IEEEabskeysecsize
    \fi\@IEEEgobbleleadPARNLSP}
\def\IEEEkeywords{\normalfont
    \if@twocolumn
      \@IEEEabskeysecsize\bfseries\textit{\IEEEkeywordsname} \relax
    \else
      \bgroup\par\addvspace{0.5\baselineskip}\centering\@IEEEabskeysecsize\textbf{\IEEEkeywordsname}\par\addvspace{0.5\baselineskip}\egroup\quotation\@IEEEabskeysecsize
    \fi\@IEEEgobbleleadPARNLSP}
\makeatother

\usepackage{amsmath,amssymb,amsfonts}
\usepackage{graphicx}
\usepackage{booktabs}
\usepackage{cite}
\usepackage[hidelinks]{hyperref}

\graphicspath{{figures/}}

\begin{document}

\title{Your Query Is My Key: Halving Attention Parameters via
Cross-Head Cyclic Rings}

\author{Sidharth~SS\\
\thanks{Independent Researcher. ORCID: 0009-0007-4442-156X.}}

\markboth{Cross-Head Cyclic Rings}{SS: Your Query Is My Key}

\maketitle

\begin{abstract}
In multi-head self-attention, the logits depend solely on the
bilinear product $U_h = W_{Q,h}W_{K,h}^{\top}$. This creates a
$d_h^{2}$-dimensional gauge redundancy in each head. Thus, we ask:
Can we remove half the query-key parameters without losing
directional attention? If we keep $W_{Q,h} = W_{K,h}$, then it will
remove half the parameters, but it will make the score matrix a
symmetric positive semidefinite Gram matrix, which cannot express
directional retrieval or negative self-scores. In this paper, we
first derive the exact identity that comes with this redundancy, and
thereafter, we describe the symmetry and sign constraints when we do
direct tying. Then, we introduce a cross-head cyclic construction,
which utilises only one projection per head and borrows the other
from a neighbouring head to calculate the scores, forming a cyclic
ring. It will reduce the query-key parameter count, GEMM count, and
optimizer states by half. In the experimental part, we have covered
associative recall, character-level language modelling, memory use,
and some small-scale expressiveness tests. In these tested cases,
the cyclic model achieved competitive performance with half the
query-key parameters compared to a paired setting. These
measurements do not establish any general advantage, but they have
shown that the cyclic restriction can retain strong performance at
the tested scale.
\end{abstract}


\section{Introduction}
\label{sec:intro}

\IEEEPARstart{S}{ince} the Transformer was proposed
\cite{vaswani2017}, self-attention has become the standard way to
mix information across a sequence. In multi-head attention, every
head keeps its own query projection and its own key projection, and
the score between two tokens is a bilinear form built from these two
matrices. This design is flexible, but every head then has its own
pair of learned matrices, and during autoregressive decoding the keys
and values of every past token must be stored: the key-value cache
grows with the sequence length and dominates the memory used at
inference. Multi-query attention \cite{shazeer2019}, grouped-query
attention \cite{ainslie2023}, and multi-head latent attention
\cite{deepseek2024} were introduced to shrink this cache by sharing or
compressing the key and value projections.

Our paper starts from a simple observation. The query projection and
the key projection do not enter the score separately; only their
product enters it. So the pair $(W_Q, W_K)$ carries a continuous
redundancy: many different pairs represent exactly the same function.
We derive the exact gradient identity that belongs to this redundancy,
we describe what direct tying (setting $W_Q = W_K$) really costs, and
we then propose a construction that keeps the parameter saving without
paying the full price of tying. The construction is a \emph{cross-head
cyclic ring}: each head owns one projection matrix, and the score of
each head is built from its own projection and the projection of its
neighbour, with indices taken modulo the number of heads. One GEMM per
head replaces two, the query-key parameter count is halved, and the
query-key optimizer state is halved as well.

The redundancy we use is not new. Wang and Wang
\cite{wang2026a,wang2026b} have already characterised the gauge
symmetry of transformer attention in full generality, and they also
study the geometry of gradients on the resulting quotient. What we add
is an exact, entry-wise matrix form of the corresponding gradient
relation for the no-position case, derived from elementary rules, and
we use it as the starting point for the tying question and for the
cyclic construction. Recent studies also show that one of the three
$Q$, $K$, $V$ matrices can be removed or shared
\cite{karbevski2025,kayyam2026,kowsher2025,cordonnier2020}, but none
of them arranges one projection per head in a ring. On the empirical
side, we follow published task definitions for associative recall
\cite{arora2024} and for expressiveness testing
\cite{deletang2023,butoi2025,li2025} instead of inventing our own
benchmarks. The full experiment results and the implementation
are available at \url{https://github.com/sidhu2690/Cyclic-Ring-Attention/}.

The contributions of this paper are as follows.
\begin{itemize}
\item We derive the exact gradient identity
      $W_Q^{\top} G_{W_Q} = G_{W_K}^{\top} W_K$ for one attention
      head, and we verify it numerically in float64 (maximum
      discrepancy $1.421\times10^{-14}$ for a general upstream
      gradient, $9.992\times10^{-16}$ under cross-entropy, and at
      every step of a toy training loop).
\item We describe the two walls of direct tying: the asymmetry wall
      and the sign wall, and we note that a fixed skew-symmetric
      bridge is only a partial repair.
\item We introduce the cross-head cyclic ring, which halves the
      query-key parameter count, the GEMM count, and the optimizer
      state, and we add FastCycle, a parameter-free gate that speeds
      the ring's escape from the balanced saddle without adding a
      single parameter.
\item We report measurements on associative recall, a character-level
      language model, a memory microbenchmark, a formal
      expressiveness battery, and a multitask suite (AG News,
      CIFAR-10, and MQAR). The results are descriptive and are
      reported with their scope, as explained in
      Section~\ref{sec:scope}.
\end{itemize}

The rest of the paper is organised as follows. Section~\ref{sec:lit}
reviews the literature. Section~\ref{sec:sine} starts the technical
development from a scalar laboratory, the sine chain, and
Sections~\ref{sec:head}--\ref{sec:ring} carry the same ideas to the
attention head and to the cyclic ring, and Section~\ref{sec:lag}
explains when the ring's optimisation lags the pair and when it
cannot, and Section~\ref{sec:fast} gives FastCycle, a
parameter-free gate that speeds up that escape.
Section~\ref{sec:compare}
comments on the comparison methodology, Section~\ref{sec:measure}
collects the measurements in a continuous narrative,
Section~\ref{sec:scope} states the scope and the limitations, and
Section~\ref{sec:concl} concludes.

\section{Literature Review}
\label{sec:lit}

\subsection{Tying and Sharing in Attention}

Weight tying is a long-standing way to reduce parameters. Press and
Wolf \cite{press2017} tied the input and output embeddings of
language models, and showed that this can reduce the size of neural
translation models to less than half without harming performance.
Inside attention, the widely used sharing schemes target the key and
value side. Multi-query
attention \cite{shazeer2019} lets all heads share one set of keys and
values, and grouped-query attention \cite{ainslie2023} generalises
this to a small number of key-value groups. Multi-head latent
attention \cite{deepseek2024} instead compresses keys and values into
a low-dimensional latent vector. These three schemes reduce the
key-value parameters and the cache, but the query side of every head
stays independent.

Sharing on the query-key side has also been studied, but less
thoroughly. Cordonnier \emph{et al.}~\cite{cordonnier2020} proposed
collaborative multi-head attention, where heads learn shared
key/query dimensions and the shared dimension can be reduced by about
four times at the same accuracy. Kowsher \emph{et al.}~\cite{kowsher2025}
replaced all three matrices by one shared matrix with diagonal role
scales, which cuts the attention-block parameters by about
$66\%$. Karbevski and Mijoski \cite{karbevski2025} proved that one of
the query, key, or value matrices can be replaced by the identity
under mild assumptions, which removes $25\%$ of the attention
parameters. Kayyam \emph{et al.}~\cite{kayyam2026} then made a
systematic study of the projection-sharing variants. They report that
sharing key with value keeps quality, while sharing query with key
\emph{breaks attention directionality}, because the score matrix
becomes symmetric. This last finding is exactly the asymmetry wall we
formalise in Section~\ref{sec:walls}. None of this line of work uses
a cyclic arrangement of one projection per head, which is the
construction we propose.

\subsection{Symmetries and the Gradient Identity}

The fact that the loss sees $W_Q$ and $W_K$ only through the product
$U = W_Q W_K^{\top}$ is an instance of a gauge symmetry. Wang and
Wang \cite{wang2026a} gave a complete characterisation of the gauge
group for canonical transformer architectures, including the RoPE
case, and they showed that valid gauge transformations preserve the
model outputs to machine precision. In their companion paper, Wang
and Wang \cite{wang2026b} developed the fiber-bundle geometry of
transformers, established an Ehresmann connection on the quotient,
and proposed a gauge-aware gradient split as a practical diagnostic.
The gradient identity we derive in Section~\ref{sec:identity} is the
explicit $k\times k$ matrix form of the corresponding tangent
orthogonality for the no-position case. We derive it from the basic
rules of matrix calculus so that the relation can be checked line by
line, and we use it to motivate both the direct-tying question and
the cyclic construction. To our knowledge, the exact identity in this
elementary form has not been written down before.

\subsection{Associative Recall and Grokking-Like Transitions}

The multi-query associative recall (MQAR) task used in
Section~\ref{sec:mqar} comes from Zoology \cite{arora2024}, where it
was introduced as a formalisation of in-context recall and used to
compare attention with gated-convolution models. MQAR shows a
grokking-like transition: accuracy stays low for a long stretch of
training and then rises quickly. This behaviour resembles
grokking \cite{power2022}, and it means that a fixed training budget
and a single seed can sit on either side of the transition. We
therefore state the training regime with every number, and we do not
treat a low fixed-budget accuracy as evidence about representational
limits.

\subsection{Expressiveness Diagnostics}

The expressiveness battery in Section~\ref{sec:express} is built from
published criteria. Del\'etang \emph{et al.}~\cite{deletang2023}
trained recognizers for languages across the Chomsky hierarchy and
found that plain RNNs and transformers fail to generalise on
non-regular tasks, while models augmented with structured memory such as a stack or tape
succeed on context-free and context-sensitive tasks. Ebrahimi
\emph{et al.}~\cite{ebrahimi2020} studied the recognition of Dyck-$n$
languages with self-attention networks. Butoi \emph{et al.}~\cite{butoi2025}
turned this line of study into a proper benchmark (FLaRe), training
networks directly as recognizers with a controlled protocol. On the structured side,
ListOps \cite{nangia2018} is used as in Long Range Arena
\cite{tay2021}. Finally, the length-generalisation probe follows Li
and Cotterell \cite{li2025}, who characterised fixed-precision
soft-attention transformers with no positional encodings and showed
empirically that languages inside the characterised class generalise
across lengths while languages outside it do not. We reuse these
protocols rather than inventing new ones, so that our numbers can be
read against the published results.

\subsection{Summary}

In short, the literature gives us (i) efficient attention variants
that share key and value across heads, (ii) several ways to remove or
share individual $Q$/$K$/$V$ matrices with a known directionality
cost on the query-key side, (iii) a complete gauge-symmetry
characterisation with gradient-level diagnostics, and (iv) standard
tasks and protocols for recall and expressiveness. What is missing is
a construction that halves the query-key parameters and keeps the scores
directional.
The cyclic ring of Section~\ref{sec:ring} is our answer to this gap.

\section{The Sine Chain: A Scalar Laboratory}
\label{sec:sine}

\begin{figure}[!t]
  \centering
  \includegraphics[width=\columnwidth]{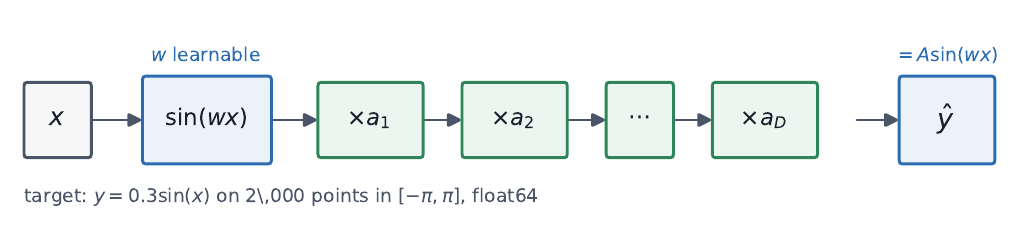}
  \caption{The sine chain. One fixed sine node with a learnable
  frequency $w$, then $D$ identity functions with learnable scales
  $a_i$. Depth grows; width never does.}
  \label{fig:sine}
\end{figure}

The study began with the sine-chain experiment, a scalar model that
is designed to isolate how gradients behave as depth increases. Its
width is one; only the number of learnable identity scales grows. This
setting makes the gradient relationships directly measurable, without
the additional structure of a large network.

The model is a fixed sine function with a learnable frequency $w$,
followed by $D$ identity functions with learnable scales $a_i$. Its
prediction is
\begin{equation}
  \hat y = a_1 a_2 \cdots a_D \sin(wx) = A \sin(wx),
  \qquad A = \prod_{i=1}^{D} a_i .
  \label{eq:sine}
\end{equation}
The target is $y = 0.3\sin(x)$ on 2{,}000 points in $[-\pi, \pi]$,
with a fixed seed and float64 arithmetic throughout
(Fig.~\ref{fig:sine}). Float64 arithmetic keeps numerical noise below
the observed gradient effects.

The float64 experiment produced two exact observations. With equal
initialisation ($a_i = 0.9$ for all $i$), the gradient of every scale
stays identical to the last printed digit at every logged step; the
gradient deviation is $0.000\times10^{0}$ at steps $0$, $1$, $2$,
$10$, $100$, $1000$, $4000$, and $7999$ of an $8000$-step run. And
the whole model behaves like a single amplitude $A$ in front of the
sine: the product of the scales is one shared degree of freedom, and
the ratios of the scales are directions the loss cannot see. The
family of points $\{a_i : \prod_i a_i = A\}$ is the first example of a
\emph{fibre}, the part of parameter space that the loss cannot
distinguish. When the initialisation is unequal, the individual
gradients differ, but their scale-weighted values remain equal. For
each data point, with $g = \partial L/\partial \hat y$ and
$A = \prod_{j=1}^{D} a_j$,
\begin{equation}
  G_{a_i} = g\, \frac{A}{a_i} \sin(wx),
  \qquad
  a_i G_{a_i} = g A \sin(wx) = a_j G_{a_j}.
  \label{eq:scalegrad}
\end{equation}
For a batch, these equalities hold after summing the per-example
contributions. The frequency gradient is a different quantity,
\begin{equation}
  G_w = \sum_t g_t A x_t \cos(wx_t),
  \label{eq:wgrad}
\end{equation}
which explains the cosine term but does not belong in $a_i G_{a_i}$.
Thus $a_i G_{a_i}$ is shared by every factor, independent of the point
on the product fibre. Every identity used here and in
Section~\ref{sec:head} was checked numerically in float64.

The same product structure appears in each attention head
\cite{vaswani2017} as $U = W_Q W_K^{\top}$. The loss depends on this
composite rather than on either factor separately. The next section
derives the resulting gradient identity, states the constraints
imposed by direct tying, and examines a cyclic alternative.

\section{From Products to the Attention Head}
\label{sec:head}

\subsection{The Multiplying Case}
\label{sec:multiply}

The simplest laboratory in which the fibre can be studied exactly is
$y = f(x)\cdot a \cdot b$ with fixed $f$, loss $L(y)$, and upstream
gradient $g = \partial L/\partial y$. The gradients are
$G_a = g f b$ and $G_b = g f a$, and one line of algebra gives
\begin{equation}
  a G_a - b G_b = a\,g f b - b\,g f a = 0 .
  \label{eq:scalar}
\end{equation}
The gradient is orthogonal to the rescaling direction $(a, -b)$ for
any differentiable loss. Along the fibre $(\alpha a, b/\alpha)$ for
any nonzero $\alpha$, the function is unchanged, and the ratio of the
factors is invisible to the loss. A single quotient coordinate
$p = ab$ captures the forward function, but this is not the same as
imposing $a = b$: that restriction gives $p = a^2$ and excludes
negative products. A new parameterisation can therefore preserve the
represented function only if it uses the product coordinate, and it
need not preserve the original optimizer trajectory.

\subsection{The Attention Pair: The Full Derivation of the $G_W$
Identity}
\label{sec:identity}

For one head, let $X \in \mathbb{R}^{n\times d}$ and
$W_Q, W_K \in \mathbb{R}^{d\times k}$, where $k$ is the head
dimension. The derivation uses the following rules: (i) linearity of
the derivative; (ii) the matrix product rule
$d(AB) = (dA)B + A(dB)$; (iii) the transpose rule
$(AB)^{\top} = B^{\top}A^{\top}$; and (iv) the Frobenius inner
product $\operatorname{tr}(A^{\top}B)$ together with the cyclic
property of the trace, $\operatorname{tr}(ABC) =
\operatorname{tr}(BCA) = \operatorname{tr}(CAB)$. The input $X$ is
upstream of the parameters, so $dX = 0$ throughout. We write
$G_{(\cdot)} = \partial L/\partial (\cdot)$.

\emph{Step 1 (from $S$ to $Q$ and $K$).}
Start from $S = c\,QK^{\top}$, with $Q = XW_Q$ and $K = XW_K$. The
product rule gives $dS = c\,dQ\,K^{\top} + c\,Q\,dK^{\top}$.
Substitute into the loss differential and use the cyclic trace:
\begin{align}
  dL &= \operatorname{tr}\!\big(G_S^{\top} dS\big)
    = c\operatorname{tr}\!\big(G_S^{\top} dQ K^{\top}\big)
    + c\operatorname{tr}\!\big(G_S^{\top} Q\, dK^{\top}\big) \notag\\
    &= \operatorname{tr}\!\big((c\,G_S K)^{\top} dQ\big)
    + \operatorname{tr}\!\big((c\,G_S^{\top} Q)^{\top} dK\big).
\end{align}
Hence
\begin{equation}
  G_Q = c\,G_S K,
  \qquad
  G_K = c\,G_S^{\top} Q.
  \label{eq:gqgk}
\end{equation}

\emph{Step 2 (from $Q$ and $K$ to the weights).}
Since $dX = 0$, we have $dQ = X\,dW_Q$ and $dK = X\,dW_K$, and the
same calculation gives $G_{W_Q} = X^{\top}G_Q$ and
$G_{W_K} = X^{\top}G_K$. Substituting \eqref{eq:gqgk},
\begin{equation}
  G_{W_Q} = c\,X^{\top} G_S K,
  \qquad
  G_{W_K} = c\,X^{\top} G_S^{\top} Q.
  \label{eq:gweights}
\end{equation}

\emph{Step 3 (the $U$ detour, where the redundancy hides).}
Set $U = W_Q W_K^{\top}$, so $S = c\,XUX^{\top}$; the loss sees $S$,
so it sees $U$ and nothing else. With $G_U = \partial L/\partial U$,
the product rule with $dX = 0$ gives $dS = c\,X\,dU\,X^{\top}$, and
the same trace argument as before yields
\begin{equation}
  G_U = c\,X^{\top} G_S X.
  \label{eq:gu}
\end{equation}
Now differentiate $U$ itself, $dU = dW_Q W_K^{\top} +
W_Q\,dW_K^{\top}$, substitute into the loss differential, and use the
cyclic trace once more:
\begin{equation}
  G_{W_Q} = G_U W_K,
  \qquad
  G_{W_K} = G_U^{\top} W_Q.
  \label{eq:gu2weights}
\end{equation}

\emph{Step 4 (close the identity).}
Multiply the first equation of Step~3 on the left by $W_Q^{\top}$:
$W_Q^{\top} G_{W_Q} = W_Q^{\top} G_U W_K$. Take the transpose of the
second equation, applying the transpose rule twice:
$G_{W_K}^{\top} = (G_U^{\top} W_Q)^{\top} = W_Q^{\top} G_U$. Multiply
both sides on the right by $W_K$:
$G_{W_K}^{\top} W_K = W_Q^{\top} G_U W_K$. The right-hand sides are
identical, therefore, for every $X$, every $(W_Q, W_K)$, and every
upstream $G_S$,
\begin{equation}
  \boxed{\;W_Q^{\top} G_{W_Q} = G_{W_K}^{\top} W_K\;}
  \label{eq:identity}
\end{equation}
\emph{Rule used: transpose of a product, applied twice.}

\emph{Reading the identity.}
Without positional mixing, the scalar rescaling
$(W_Q, W_K) \mapsto (\alpha W_Q, \alpha^{-1} W_K)$ for $\alpha > 0$
extends to the matrix-valued symmetry
$(W_Q, W_K) \mapsto (W_Q A, W_K A^{-\top})$ for invertible
$A \in GL(k)$. This preserves $U = W_Q W_K^{\top}$, and therefore $S$
and the loss. For full-column-rank factors, the transformations form
a $k^2$-dimensional gauge orbit. The boxed $k\times k$ identity
expresses tangent orthogonality to this orbit; it holds for every
upstream gradient, without assumptions on the loss or the data path.
Wang and Wang \cite{wang2026a} characterise this query-key gauge
symmetry and show that RoPE restricts the allowed transformations;
differentiating the no-position product-preserving symmetry gives the
tangent gradient identity derived here. Numerical checks in float64
gave a maximum discrepancy of $1.421\times10^{-14}$ for the matrix
identity, and $9.992\times10^{-16}$ under cross-entropy. The identity
was also checked at every step of a toy training loop.

A natural thought follows: if the loss only sees $U$, then tie the
query and key projections to one matrix per head, shared across both
roles, reducing the query-key parameters and their optimizer state by
half. This thought is true for the scalar case, and it is half true
for matrices; the second half is where the walls live.

\subsection{One Matrix, Two Walls}
\label{sec:walls}

Let $W$ be a $d\times k$ matrix, let $Z = XW$, and let the score
between token $i$ and token $j$ be $s_{ij} = c\, z_i \cdot z_j$.
Three properties of the score matrix are not accidents:

\begin{enumerate}
\item \emph{Property 1.} $S$ is symmetric, because the dot product is
      commutative. In attention language: token $i$ finds token $j$
      interesting exactly as much as token $j$ finds token $i$
      interesting, with the same signed score.
\item \emph{Property 2.} The diagonal $s_{ii} = c\|z_i\|^2 \ge 0$; a
      token's self-similarity cannot be negative.
\item \emph{Property 3.} Every weighted energy
      $\sum_{i,j} w_i w_j s_{ij} = c\|\sum_i w_i z_i\|^2 \ge 0$; for
      $c > 0$, $S$ is positive semi-definite.
\end{enumerate}

Property~1 is the \emph{asymmetry wall}: a retrieval circuit in a
language model is directional by construction, and a plain tied score
cannot express direction. Properties~2 and~3 are the \emph{sign
wall}: a plain tied model cannot produce a negative self-score or
arbitrary signed interactions. A partial repair is possible:
inserting a fixed skew-symmetric matrix $J$ between the tied
projections gives $S = c\,ZJZ^{\top}$, which can produce signed,
generally nonsymmetric scores and no longer has the
positive-semidefinite Gram property, although the score bilinear form
remains confined to the subspace spanned by the columns of $W$. A
different route replaces one projection by the identity matrix
outright: the score then depends on a single learned matrix instead
of a tied product, which avoids the Gram form altogether at the same
query-key saving \cite{karbevski2025}. Kayyam \emph{et al.}~\cite{kayyam2026} report the
same directionality loss empirically for the shared-query-key
variant. The cyclic construction below instead uses neighbouring head
projections to introduce cross-head directionality. These changes
alter the function family, rather than removing a mere implementation
constraint.

\section{The Ring: One Matrix per Head, Shared Around a Ring}
\label{sec:ring}

\begin{figure*}[!t]
  \centering
  \includegraphics[width=0.9\textwidth]{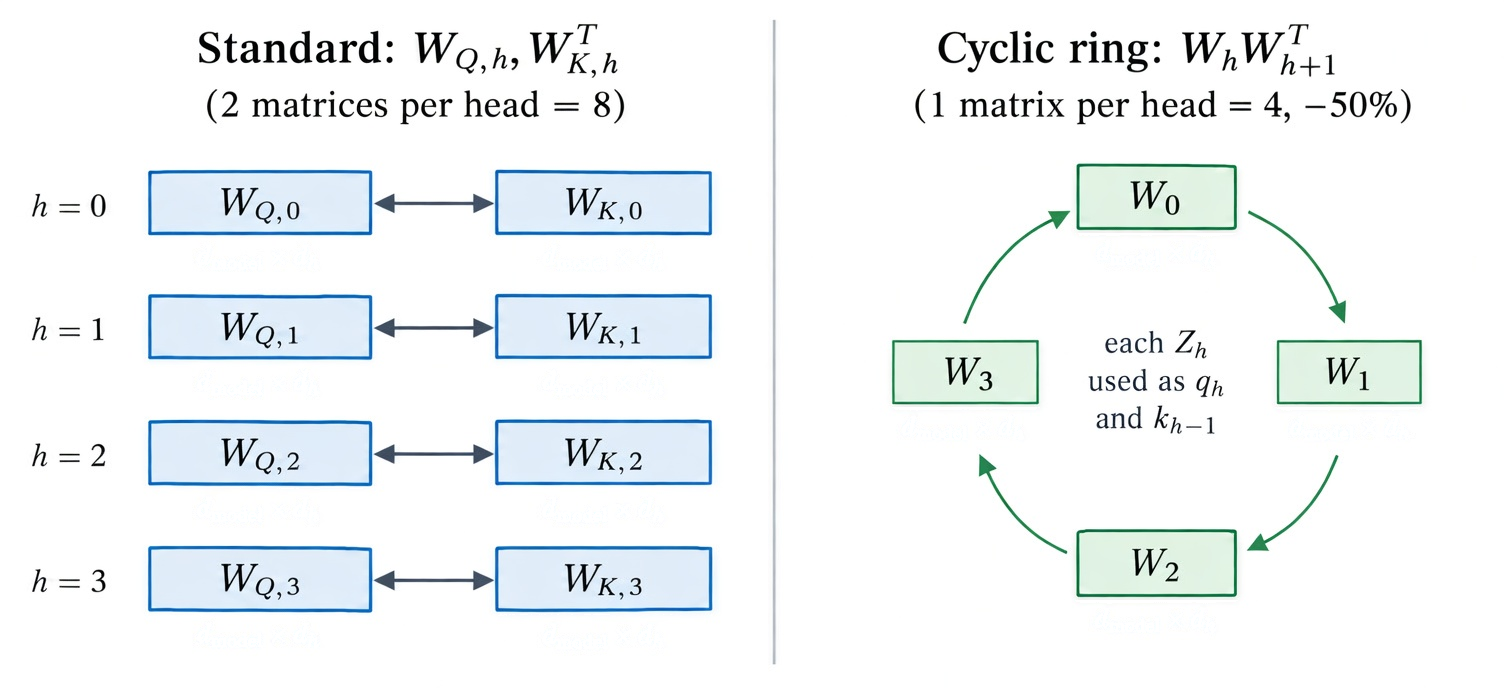}
  \caption{Standard query-key parameterisation versus the cyclic ring
  at $H = 4$. Left: standard attention uses two separate matrices per
  head to form $W_{Q,h}W_{K,h}^{\top}$ ($2H = 8$ matrices). Right:
  cyclic attention arranges one matrix per head in a ring, so each
  head forms $W_h W_{(h+1)\bmod H}^{\top}$ ($H = 4$ matrices, cutting
  query-key parameters by $50\%$).}
  \label{fig:ring}
\end{figure*}

The cyclic construction keeps the parameter saving and introduces
directionality through sharing between heads, rather than tying the
same matrix within each head. Head $h$ owns one $d\times d_h$ matrix
$W_h$ and computes $Z_h = XW_h$ once. The scores for head $h$ are
built from its own projection and the projection of its neighbour, as
shown in Fig.~\ref{fig:ring}:
\begin{equation}
  S_h = \frac{1}{\sqrt{d_h}}\, R(Z_h)\, R(Z_{h+1})^{\top},
  \qquad h = 0, \ldots, H-1,
  \label{eq:ring}
\end{equation}
with indices taken modulo $H$, and $R$ the split-half RoPE of
\cite{su2021}. Each $Z_h$ is computed once and used twice: as the
query for head $h$, and as the key for head $h-1$. At $H = 2$, the
two heads form a reciprocal pair of bilinear forms rather than a
longer cycle. Note that this ``ring'' denotes the cyclic head
wiring of Fig.~\ref{fig:ring}; Ring Attention \cite{liu2024} is a
distributed context-parallel scheme and is unrelated.

What the ring keeps and what it pays: the query-key projection count
is halved (one GEMM per head instead of two), the query-key parameter
count is halved, the query-key optimizer state is halved, and the
function family is smaller: the pair's full family of bilinear forms
is traded for the ring's structured one. This is a change to the
model family, not a reparameterisation of the pair. What the ring does
not do is recover the pair's missing directions, and it should not be
expected to.

\subsection{When the Ring Lags and When It Does Not}
\label{sec:lag}

The measurements in Section~\ref{sec:measure} show a repeated
pattern: on sparse associative-recall probes (MQAR), the cyclic
model lifts off later than the pair, with the phase-transition
crossover about 540 steps later on average (mean crossing step 2171
against 1632 for the pair), whereas on dense multi-feature tasks
(AG News, CIFAR-10, and character-level language modelling) the
delay vanishes. This difference is governed by how long the network
remains trapped near the initial $Z_H$-symmetric saddle, where $H$
heads compete for a single sparse retrieval mode ($M = 1$ circuit
$(u,v,\Delta)$ with $u \perp v$), before role specialization breaks the symmetry around
the ring. Three mechanisms slow the ring while it sits at this
saddle.

\emph{Optimizer dilution.} In the pair, the query matrix of head $h$
steps along its own gradient $G_h^{Q}$ and the key matrix along
$G_h^{K}$, without interference. In the ring, the single matrix
$W_h$ receives both role gradients, $G_h^{Q} + G_{h-1}^{K}$, and one
RMS-normalised step must serve the sum. Let
$r_h = \|G_{h-1}^{K}\|_F / \|G_h^{Q}\|_F$ be the ratio of the two
role gradients at node $h$. When the two are uncorrelated in
expectation, the fraction of the pair's combined role progress that
survives is
\begin{equation}
\begin{aligned}
d(r_h) &= \frac{\|G_h^{Q} + G_{h-1}^{K}\|_F}
               {\|G_h^{Q}\|_F + \|G_{h-1}^{K}\|_F} \\
       &= \frac{\sqrt{1 + r_h^{2}}}{1 + r_h},
\end{aligned}
\label{eq:dilution}
\end{equation}
which is symmetric under $r_h \leftrightarrow 1/r_h$, reaches its
minimum $1/\sqrt{2} \approx 0.707$ at $r_h = 1$, and returns to $1$
as $r_h$ tends to $0$ or $\infty$. Because the ring score is
bilinear in two matrices that both grow from small initialisation at
rate $d$, the operator $U_h$ grows at rate $d^{2}$; at the balanced
point $r_h = 1$ this is exactly half the pair's rate. Whenever the
two roles carry comparable gradient mass, the ring's logits open at
half speed.

\emph{The ghost terms.} Second, one shared matrix loads both roles.
If the task is dominated by a single association
$U^{\ast} \propto uv^{\top}$ with $u \perp v$, every head receives
query signal along $u$ and key signal along $v$, so
$W_h = u\alpha_h^{\top} + v\beta_h^{\top}$ carries both directions,
where $\alpha_h$ serves head $h$ as query and $\beta_h$ serves head
$h-1$ as key. The pair keeps the two directions in separate matrices
and produces only the desired term. The ring expands into four:
\begin{equation}
\begin{aligned}
U_h(\Delta) &= (\alpha_h^{\top} R_{\Delta} \beta_{h+1})\, u v^{\top}
  + (\alpha_h^{\top} R_{\Delta} \alpha_{h+1})\, u u^{\top}\\
&\quad + (\beta_h^{\top} R_{\Delta} \beta_{h+1})\, v v^{\top}
  + (\beta_h^{\top} R_{\Delta} \alpha_{h+1})\, v u^{\top},
\end{aligned}
\label{eq:ghost}
\end{equation}
where the first term is the desired circuit and the other three are
ghosts. At the symmetric saddle, where neighbouring heads are still
statistically identical, the four coefficients have equal magnitude,
and because $\{uv^{\top}, uu^{\top}, vv^{\top}, vu^{\top}\}$ are
orthogonal for $u \perp v$, the three unwanted terms carry
$\sqrt{3}/2 \approx 86.6\%$ of the kernel's Frobenius norm. To
escape, $\alpha_h$ and $\beta_{h+1}$ must share one rotation plane
of $R_{\Delta}$ while $\beta_{h+1}$ and $\alpha_{h+1}$ occupy
disjoint planes: the ring must break its own cyclic symmetry $Z_H$
before it can form a clean rank-1 circuit. The pair never waits; its
cross-terms are zero by construction.

\emph{Idle neighbours.} Third, at large $H$ the waking head drags
idle neighbours. With $M = 1$, each head starts with an independent
random alignment $z_h(0)$ with the target circuit (standard
deviation $\sigma$), and softmax feedback amplifies the best
initial draw, so the winner starts ahead by the extreme-value margin
$\approx \sigma\sqrt{2\ln H}$. In the pair the winner's two matrices
are private and the other $H-1$ heads cannot touch them. In the ring
the winner $h^{\ast}$ shares $W_{h^{\ast}}$ and
$W_{h^{\ast}+1}$ with its two neighbours; with high probability those
neighbours are still idle, so their unspecialised gradients dilute
the winner's steps, and every update toward $(u, v)$ writes ghost
terms into the neighbours, which push back. The penalty sits exactly
at the active-to-idle boundaries of the ring.

\emph{When the lag vanishes, and when capacity bites.} The three
mechanisms above operate only while neighbouring heads remain
locked in the unbroken $Z_H$-symmetric saddle on a sparse retrieval
objective. When a single layer is driven by $M \approx H$
orthogonal modes simultaneously, or when the model trains on dense,
heterogeneous data (such as AG News, CIFAR-10, and real-text
language modelling), distinct gradients break the cyclic symmetry
across heads from the earliest steps; although both roles of each
shared matrix still carry comparable gradient norms ($r_h \approx
1$, $d(r_h) \approx 1/\sqrt{2}$), the ghost cross-terms decouple
immediately and the liftoff lag disappears. Conversely, two
extremes remain hard for the ring: (i) at very large head counts on
sparse single-circuit recall ($H \in \{16, 32\}$ in
Table~\ref{tab:mqar}), idle-neighbour drag and saddle dilution can
stall liftoff within a fixed step budget, and (ii) when multiple
exact-match retrieval circuits are packed into a narrow per-head
dimension ($d_h = 16$), the ring's 50\% reduction in query--key
parameters becomes a genuine subspace-capacity bottleneck rather
than a mere liftoff delay.

\subsection{FastCycle Attention}
\label{sec:fast}

The analysis above suggests a direct fix. When the ring sits in the
balanced saddle ($r_h \approx 1$), the shared matrix receives a
compromised step; if that step were amplified while the ring is
balanced, and withdrawn once the two roles separate, the ring could
leave the saddle faster without changing its parameter count.
FastCycle attention (called \texttt{FastCyclicAttention} in the
implementation) is exactly this gate on top of the plain ring, and
it adds no learnable parameters.

Each step, the gate measures the role-gradient imbalance across
heads, $\delta = \mathrm{RMS}_h\!\big(\ln \|G_h^{K}\|_F /
\|G_h^{Q}\|_F\big)$, which is the log form of the ratio $r_h$ from
Section~\ref{sec:lag}. The signal is smoothed by an exponential
moving average and latched to its running maximum after a short
warmup (steps 5 to 20), then mapped through a fixed Gaussian,
$\gamma = 1 + w(t)\,(\gamma_{\max}-1)\exp(-s^{2}/2\sigma^{2})$
with $\gamma_{\max} = 1.55$ and $\sigma = 0.35$, using asymmetric
smoothing so that $\gamma$ rises quickly and falls slowly. The gated
weight is the anchor extrapolation
\begin{equation}
  W_{\mathrm{eff}} = W_{0} + \gamma\,(W - W_{0}),
  \label{eq:fast}
\end{equation}
where $W_{0}$ is the initial matrix and gradients pass through at
full strength (straight-through). While the ring is balanced,
$\delta \approx 0$ and $\gamma$ rises towards $1.55$, so the
effective move is stretched by up to $55\%$, exactly the
direction the dilution factor of \eqref{eq:dilution} says the ring
needs. As soon as the two roles differentiate, $\delta$ grows,
$\gamma$ falls back to $1$, and the module is exactly the plain
ring again. In the measurements, FastCycle closes part of the
liftoff gap (mean crossing step 1867 against 2171 for the plain
ring and 1632 for the pair) and matches the other arms on the
real-world tasks, with the same parameter, optimizer-state, and
GEMM counts as the plain ring.

\section{On Comparing Two Different Paths}
\label{sec:compare}

The three arms use different parameterisations and
therefore follow different optimisation paths. Their results are
reported as measurements under the stated budgets, not as a strict
like-for-like comparison. Several settings show similar performance across the three
arms; some length-generalisation cases show substantial
variation. The experiments do not establish general superiority for
either parameterisation. They describe the measured trade-offs of the
cyclic restriction at this scale.

\section{What the Measurements Say}
\label{sec:measure}

Every experiment below uses the same base configuration unless stated
otherwise: a 2-layer LLaMA-style block \cite{touvron2023} with
RMSNorm \cite{zhang2019}, SwiGLU \cite{shazeer2020}, and split-half
RoPE \cite{su2021}, head dimension 16, Adam \cite{kingma2015}, batch
128, and seeds $\{0, 1, 2\}$ with means and sample standard
deviations reported; every experiment runs three arms: pair,
cyclic, and FastCycle. All numbers are from runs on $2\times$RTX A4000
GPUs. The implementation used for these experiments is available at
\url{https://github.com/sidhu2690/Cyclic-Ring-Attention/}.

\subsection{The MQAR Stress Test}
\label{sec:mqar}

\begin{table}[!t]
  \centering
  \caption{MQAR exact-answer accuracy, mean$\,\pm\,$sample SD over
  seeds $\{0,1,2\}$. Chance $= 0.004$. Regimes: A $=$ 4000 steps,
  lr $2\times10^{-3}$; C $=$ 8000, $7\times10^{-4}$; H $=$ 8000,
  $5\times10^{-4}$; L $=$ 30{,}000, $7\times10^{-4}$. fast $=$
  FastCycle (Section~\ref{sec:fast}).}
  \label{tab:mqar}
  \footnotesize
  \begin{tabular}{ccccccc}
    \toprule
    S & D & H & reg & pair & cyclic & fast \\
    \midrule
    8  & 64  & 4  & C & $.960\pm.003$ & $.956\pm.007$ & $.952\pm.008$ \\
    8  & 128 & 8  & C & $.963\pm.006$ & $.949\pm.005$ & $.955\pm.004$ \\
    8  & 256 & 16 & C & $.437\pm.420$ & $.085\pm.007$ & $.084\pm.003$ \\
    8  & 256 & 16 & L & $.972\pm.013$ & $.083\pm.004$ & $.086\pm.001$ \\
    8  & 512 & 32 & C & $.089\pm.002$ & $.091\pm.006$ & $.072\pm.007$ \\
    8  & 512 & 32 & H & $.102\pm.007$ & $.116\pm.014$ & $.110\pm.021$ \\
    8  & 512 & 32 & L & $.946\pm.016$ & $.379\pm.354$ & $.131\pm.040$ \\
    10 & 64  & 4  & A & $.937\pm.007$ & $.039\pm.003$ & $.040\pm.004$ \\
    10 & 64  & 4  & C & $.950\pm.010$ & $.924\pm.048$ & $.957\pm.003$ \\
    10 & 128 & 8  & C & $.964\pm.002$ & $.944\pm.008$ & $.946\pm.005$ \\
    10 & 256 & 16 & C & $.482\pm.391$ & $.073\pm.002$ & $.070\pm.003$ \\
    10 & 256 & 16 & L & $.966\pm.001$ & $.073\pm.006$ & $.068\pm.003$ \\
    10 & 512 & 32 & C & $.056\pm.003$ & $.070\pm.001$ & $.054\pm.001$ \\
    10 & 512 & 32 & H & $.073\pm.007$ & $.083\pm.013$ & $.077\pm.005$ \\
    10 & 512 & 32 & L & $.054\pm.005$ & $.639\pm.456$ & $.062\pm.005$ \\
    \bottomrule
  \end{tabular}
\end{table}

\begin{figure}[!t]
  \centering
  \includegraphics[width=\columnwidth]{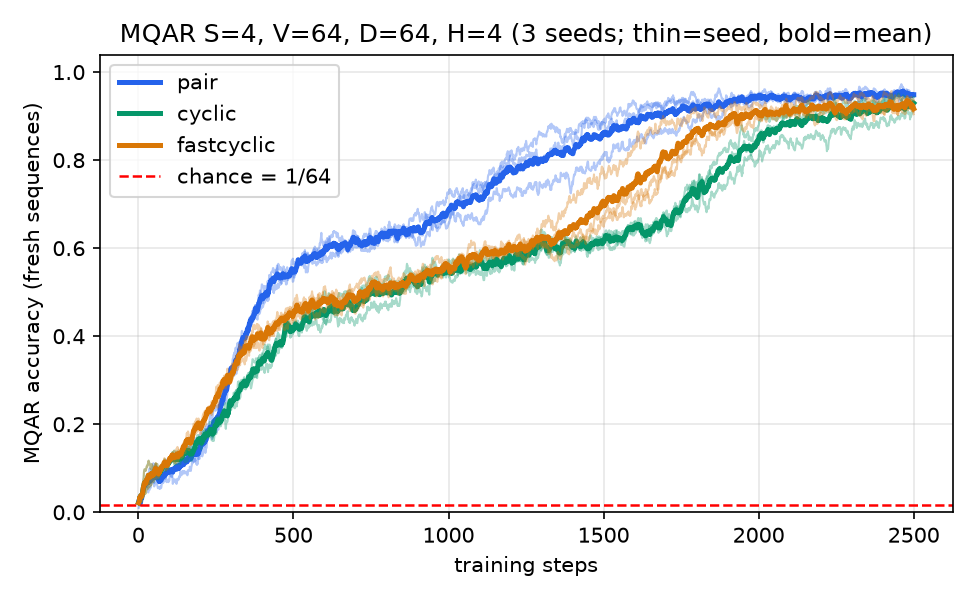}
  \caption{MQAR accuracy during training in a scaled-down reference
  run (S$=$4, V$=$64, D$=$64, H$=$4, batch 64, lr $7\times10^{-4}$,
  2500 steps, 3 seeds; from the repository). The arms stay low for
  about 1500 steps, then cross $0.9$ at mean step 1632 (pair), 1867
  (FastCycle), and 2171 (cyclic), and finish between $.90$ and
  $.95$. Chance is $1/64$.}
  \label{fig:grok}
\end{figure}

The task is a Zoology-style multi-query associative recall (MQAR)
\cite{arora2024}: a sequence of key--value pairs
$k_1 v_1 \ldots k_S v_S$ followed by a query $Q$, and the target is
the value paired with the key equal to $Q$ (chance $= 1/256$). This
task directly probes the query-key retrieval pathway; the full
three-arm matrix is in Table~\ref{tab:mqar}. Four training regimes
are reported: A (4000 steps, lr $2\times10^{-3}$), C (8000 steps,
lr $7\times10^{-4}$), H (8000 steps, lr $5\times10^{-4}$), and L
(30{,}000 steps, lr $7\times10^{-4}$). Under regime C the small
widths are learned by all three arms: at D $= 64$ and $128$ the
means sit between $.924$ and $.964$ with gaps of at most $.026$.
The wide configs are unstable at the 8000-step budget, and here the
pair fails too, not only the ring: at D $= 256$, H $= 16$ the pair
reaches only $.437 \pm .420$ (S $= 8$) and $.482 \pm .391$
(S $= 10$) while the ring sits near $.08$, and at D $= 512$,
H $= 32$ all three arms stay between $.05$ and $.12$. A longer
budget (regime L) separates the arms: the pair recovers at
D $= 256$ ($.972$, $.966$) while the ring stays near $.08$, the
high-redundancy weakness of Section~\ref{sec:lag}, and at
D $= 512$ the outcome flips with S (pair $.946$ against cyclic
$.379$ at S $= 8$; pair $.054$ against cyclic $.639$ at S $= 10$),
so neither wiring is reliable there. MQAR also shows a
phase transition \cite{power2022}: before the transition, accuracy
remains low and is sensitive to seed and training budget. The rise
resembles grokking, but its reason is different, so we call it a
phase transition throughout. Fig.~\ref{fig:grok} shows this behaviour in a
scaled-down reference run: the arms stay low for about 1500 steps
and then cross $0.9$ at mean step 1632 (pair), 1867 (FastCycle),
and 2171 (cyclic), finishing between $.90$ and $.95$, the delay
pattern analysed in Section~\ref{sec:lag}, a difference of speed,
not of final accuracy. Results are therefore reported with the
training regime stated, and low-accuracy cells are not treated as
evidence about representational limits. The learning rate is
another confound: at lr $= 2\times10^{-3}$ (regime A) the ring arm
fails at S $= 10$, D $= 64$, H $= 4$ ($.039$ against $.937$ for
the pair).

FastCycle tracks the plain ring closely: within $.04$ in every
8000-step cell, with its clearest win over the pair at S $= 10$,
D $= 64$, H $= 4$ under regime C ($.957$ against $.950$). At the
long budget on D $= 512$ it stays near the floor while the plain
ring partially learns (S $= 8$: $.131$ against $.379$; S $= 10$:
$.062$ against $.639$), so the gate does not help there. These
results do not establish general superiority for any arm; they map
where each wiring learns, where it stalls, and how fast it crosses
the transition.

\subsection{A Real-Text Character Language Model}
\label{sec:charlm}

The second laboratory is a small character-level LM (276\,KB corpus,
2-layer, T$=$64, batch 32), which is where the cyclic family was
first checked against held-out cross-entropy (CE, lower is better).
The original protocol used a 146-symbol vocabulary; the raw re-run
result files record vocab 65, so the parameter counts below differ
from the original run's (1{,}385{,}472 for the pair). The H $= 16$ run used four configurations: pair
(FF $= 512$), cyclic (FF $= 512$), FastCycle (FF $= 512$), and
cyclic with some query-key savings reinvested into FF $= 600$
(near-matched: 1{,}348{,}096 params against 1{,}344{,}000, a
$0.2\%$ increase). The numbers below are the full re-run shipped
in the repository.

\begin{table}[!t]
  \centering
  \caption{Character-level LM held-out cross-entropy. Entries are
  means$\,\pm\,$sample SD over 3 seeds. An asterisk marks the
  near-matched FF$=$600 setting.}
  \label{tab:charlm}
  \footnotesize
  \begin{tabular}{ccc}
    \toprule
    config & params & val CE \\
    \midrule
    16 pair          & 1{,}344{,}000 & $1.7046 \pm .0167$ \\
    16 cyclic        & 1{,}212{,}928 & $1.7145 \pm .0345$ \\
    16 fast          & 1{,}212{,}928 & $1.7091 \pm .0212$ \\
    16 cyclic FF600$^{\ast}$ & 1{,}348{,}096 & $1.7361 \pm .0098$ \\
    32 pair          & 5{,}309{,}440 & $2.2026 \pm .0268$ \\
    32 cyclic        & 4{,}785{,}152 & $2.2055 \pm .0180$ \\
    32 fast          & 4{,}785{,}152 & $2.1940 \pm .0245$ \\
    \bottomrule
  \end{tabular}
\end{table}

At H $= 16$ the four configurations have similar held-out CE across
three seeds (Table~\ref{tab:charlm}); the FF $= 600$ cyclic model is
near-matched, with $0.2\%$ more parameters than the pair, so this is
not an exact parameter-parity comparison. At H $= 32$, held-out CE
was $2.2026$ (pair), $2.2055$ (cyclic), and $2.1940$ (FastCycle).
These results are descriptive and do not establish a reliable
advantage.

\subsection{The Memory Microbenchmark}
\label{sec:memory}

The objective of the whole programme changed from time to memory, so
the measurement that matters most is the byte count. Measured on the
H$=$32/D$=$512/FF$=$1024/2-layer configuration, BS$=$16, T$=$256, one
train step (Table~\ref{tab:memory}).

\begin{table}[!t]
  \centering
  \caption{Memory microbenchmark (one step, BS$=$16, T$=$256, vocab
  256). Adam state is m$+$v for all parameters. FastCycle has the
  same counts as the plain ring (no extra parameters).}
  \label{tab:memory}
  \small
  \begin{tabular}{cccc}
    \toprule
    & pair & cyclic & FastCycle \\
    \midrule
    total params    & 5.505M  & 4.981M  & 4.981M \\
    QK params       & 1{,}048{,}576 & 524{,}288 & 524{,}288 \\
    params (fp32)   & 21.0 MiB & 19.0 MiB & 19.0 MiB \\
    Adam m$+$v      & 42.0 MiB & 38.0 MiB & 38.0 MiB \\
    \bottomrule
  \end{tabular}
\end{table}

The ring cuts query-key parameters by $50.0\%$ and total
parameters by $9.5\%$; these deltas are exact, not extrapolated.
Peak allocation was not reliably measurable in the re-run
(allocator-dependent) and is therefore not reported. Extrapolated (never measured at 7B, flagged as
extrapolation), a 7B-class model with D$=$4096/H$=$32 would carry
roughly half a billion fewer query-key parameters, about 6\,GiB
(6.44\,GB) of fp32 parameter and Adam-moment state for a 32-layer
model. This is a parameter-state estimate only; activation memory
depends on batch size, sequence length, and implementation.
Compatibility with GQA \cite{ainslie2023} or MQA \cite{shazeer2019}
was not tested.

\subsection{The Expressiveness Battery}
\label{sec:express}

A fair question at this point is whether the ring's smaller family
buys its efficiency with raw competence on anything. To answer it
with published criteria rather than home-made ones, a battery was
built from three benchmark families used in the recent
expressiveness literature for small transformers: binary
formal-language recognizers trained per language
\cite{deletang2023,butoi2025,ebrahimi2020}, ListOps \cite{nangia2018}
as used in Long Range Arena \cite{tay2021}, and a
length-generalisation probe \cite{butoi2025,li2025}. All labels come
from the task generators, with binary membership checked by the
recognizer rules. Each seed uses a deterministic held-out set
generated separately from training, balanced across classes (2048
examples for binary tasks; 204 per class for ListOps). Results are in
Table~\ref{tab:express}. Every cell of the grid was completed for
all three arms: the formal battery, both ListOps budgets, and the
full length-generalisation sweep.

\begin{table}[!t]
  \centering
  \caption{Formal battery, in-distribution (T$=$64). Chance $= .5$.
  Mean$\,\pm\,$sample SD over 3 seeds. Every measured cell is shown;
  fast $=$ FastCycle (Section~\ref{sec:fast}).}
  \label{tab:express}
  \footnotesize
  \setlength{\tabcolsep}{4pt}
  \begin{tabular}{ccccc}
    \toprule
    task & H & pair & cyclic & fast \\
    \midrule
    first  & 4,8,16,32 & 1.0000 & 1.0000 & 1.0000 \\
    parity & 4  & $.4907\pm.0082$ & $.4907\pm.0082$ & $.4907\pm.0082$ \\
           & 8  & $.4950\pm.0126$ & $.4950\pm.0126$ & $.4950\pm.0126$ \\
           & 16 & $.4969\pm.0135$ & $.4950\pm.0126$ & $.4950\pm.0126$ \\
           & 32 & $.5093\pm.0082$ & $.5093\pm.0082$ & $.5093\pm.0082$ \\
    cycle3  & 4  & $.5011\pm.0097$ & $.5013\pm.0042$ & $.5016\pm.0042$ \\
            & 8  & $.5005\pm.0043$ & $.4951\pm.0045$ & $.4956\pm.0015$ \\
            & 16 & $.4969\pm.0029$ & $.5020\pm.0034$ & $.4948\pm.0090$ \\
            & 32 & $.5042\pm.0078$ & $.5011\pm.0024$ & $.4984\pm.0066$ \\
    dyck2   & 4  & $.9990\pm.0010$ & $.9969\pm.0029$ & $.9998\pm.0003$ \\
            & 8  & $.9979\pm.0010$ & $.9997\pm.0006$ & $.9987\pm.0015$ \\
            & 16 & $.9993\pm.0006$ & $.9966\pm.0043$ & $.9995\pm.0008$ \\
            & 32 & $.9474\pm.0869$ & $.9390\pm.0678$ & $.9561\pm.0291$ \\
    \bottomrule
  \end{tabular}
\end{table}

The two regular global-count languages produce accuracies near chance
for all three arms at every tested size, consistent with the
small-transformer failure reported by Del\'etang
\emph{et al.}~\cite{deletang2023}. These measurements do not resolve
small differences between architectures. The context-free language is
learned by all three arms at every tested size (at least $.93$). ListOps (depth
$\le 4$, T $= 64$, 10-way, chance $= .1$, 8000 steps) yields H $= 4$
pair $.689$ against cyclic $.678$ against FastCycle $.683$;
H $= 8$ $.691$ / $.694$ / $.692$; H $= 16$ $.639$ / $.650$ / $.648$;
H $= 32$ $.597$ / $.601$ / $.613$. All are above chance and training
accuracy remains limited. With three seeds, these small differences
are descriptive rather than a formal equivalence result
($|\Delta| \le .016$). On the length-generalisation probe (train
T $\in \{32, 64\}$, test 32/64/128), in-distribution accuracy stays
near $1.0$ while out-of-distribution accuracy is seed-dependent and
ranges from chance (first, H $= 4$, pair $.4967$ at T $= 128$ after
train T $= 32$) to near ceiling (Dyck-2, H $= 8$), consistent with
the failure mode reported in \cite{butoi2025,li2025} at the hard
end. The three arms interleave with no consistent ordering: cyclic
is ahead on first at H $= 4$ (T $= 128$: $.731$ against pair
$.497$), while pair is ahead on Dyck-2 at H $= 16$ after train
T $= 64$ (T $= 128$: $.759$ against cyclic $.561$). Standard
deviations reach $\approx .29$ at $n = 3$, so none of these gaps is
resolved; they are logged descriptively, not as results.

\subsection{The GPU Multitask Suite}
\label{sec:multitask}

A second suite runs all three arms side by side on three tasks:
MQAR, AG News text classification, and CIFAR-10 image
classification. The suite finished without failure: 58 of 58 runs
and 29 of 29 benchmarks completed with 0 failed.
Table~\ref{tab:multitask} gives the protocol and
Table~\ref{tab:multitaskres} the final test accuracies. Unlike the
synthetic probes above, AG News and CIFAR-10 are standard
real-world tasks. On these real tasks the three arms are
statistically indistinguishable: AG News lands at $.899$ for every
arm and layer count (best gap $0.1$ percentage points), and
CIFAR-10 stays within $0.5$ points of the pair. The multitask MQAR
rows ($V = 8192$, chance $1/8192$) are far harder: at width $64/4$
the pair reaches $.227$ against $.065$ for the ring, and at
$256/16$ all three arms sit near the floor, another regime where
the plain pair keeps its speed advantage.

\begin{table}[!t]
  \centering
  \caption{GPU multitask suite: protocol and status. Pair and
  cyclic use the same optimizer and learning rate within each
  task; models train from scratch with no pretrained weights.}
  \label{tab:multitask}
  \footnotesize
  \begin{tabular}{p{0.30\columnwidth}p{0.60\columnwidth}}
    \toprule
    item & setting \\
    \midrule
    status & 58/58 runs completed, 0 failed; 29/29 \\
           & benchmarks; final accuracies in \\
           & Table~\ref{tab:multitaskres} \\
    hardware & NVIDIA RTX A4000, FP32, one process \\
             & per GPU \\
    MQAR & Adam, LR $7\times10^{-4}$, 3{,}000 steps, \\
         & batch 128, 8{,}192 independently generated \\
         & held-out examples \\
    AG News, & AdamW, peak LR $4\times10^{-4}$, warmup \\
    CIFAR-10 & $+$ cosine decay, batch 32, 800/1{,}000 \\
             & steps, official test split; monitor is \\
             & a held-out subset \\
    models & two and four layers; width/head $64/4$ and \\
           & $256/16$ ($128/16$ for the MQAR worst case) \\
    \bottomrule
  \end{tabular}
\end{table}

\begin{table}[!t]
  \centering
  \caption{Multitask suite final test accuracy (mean$\,\pm\,$sample
  SD over 3 seeds); fast $=$ FastCycle. AG News and CIFAR-10 rows
  are 2-layer and 4-layer models; MQAR rows use $V=8192$, S $= 8$,
  3000 steps (chance $1/8192$).}
  \label{tab:multitaskres}
  \footnotesize
  \setlength{\tabcolsep}{3pt}
  \begin{tabular}{ccccc}
    \toprule
    task & layers & pair & cyclic & fast \\
    \midrule
    AG News  & 2 & $.8999\pm.0026$ & $.8996\pm.0013$ & $.8998\pm.0014$ \\
    AG News  & 4 & $.8987\pm.0014$ & $.8986\pm.0021$ & $.8977\pm.0008$ \\
    CIFAR-10 & 2 & $.5287\pm.0062$ & $.5246\pm.0060$ & $.5258\pm.0066$ \\
    CIFAR-10 & 4 & $.5594\pm.0046$ & $.5557\pm.0066$ & $.5577\pm.0052$ \\
    MQAR 64/4  & 2 & $.2266\pm.0345$ & $.0650\pm.0441$ & $.1007\pm.0301$ \\
    MQAR 128/8 & 2 & $.0601\pm.0708$ & $.0038\pm.0016$ & $.0026\pm.0008$ \\
    MQAR 256/16 & 2 & $.0070\pm.0004$ & $.0060\pm.0020$ & $.0059\pm.0018$ \\
    \bottomrule
  \end{tabular}
\end{table}

\begin{figure*}[!t]
  \centering
  \includegraphics[width=\textwidth]{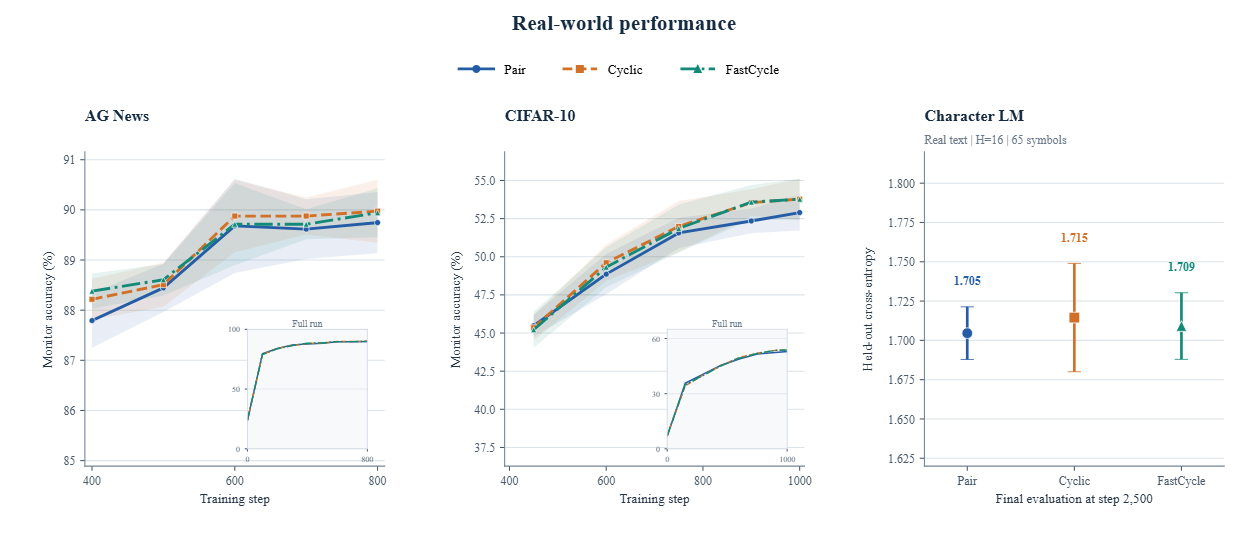}
  \caption{Real-world performance of the three arms. Left and centre:
  monitor accuracy during training on AG News and CIFAR-10 (insets:
  full runs). Right: held-out cross-entropy of the character-level LM
  at the final evaluation (step 2500; real text, H $= 16$, 65
  symbols). Shaded bands and error bars show the spread across the
  three seeds.}
  \label{fig:realworld}
\end{figure*}

\subsection{Reading the Tables Together}
\label{sec:readtogether}

The measurements show broad agreement wherever the task is
actually learned: the three arms land within noise on AG News,
within half a point on CIFAR-10, within $0.01$ cross-entropy on the
character-level LM, and at the same chance or ceiling on the formal
battery. Where they separate is speed, the phase-transition crossover,
and the high-redundancy MQAR cells, where the pair is faster or
more likely to converge. These results describe performance at the
tested scale and do not establish a general ranking.

\section{Scope and Limitations}
\label{sec:scope}

The following limitations define the scope of the results.

\emph{First}, we do not know whether the cyclic model is a restricted
function family or not. The pair uses independent query and key
projections, and the cyclic model constrains those projections to
neighbouring matrices, but with more training the cyclic model
attains similar results, the phase-transition problem is present in
both, and the two perform similarly on harder tasks, which is where
transformers, especially multi-head transformers, are used in the
real world. The experiments do not establish equivalence between
the two families; they measure performance for the tested tasks and
budgets only.

\emph{Second}, optimisation speed varies by setting. In the paired
trajectories used to characterise the phase transition, the cyclic model
crossed the transition about 540 steps later on average (2171
against 1632), and FastCycle sits between them (1867).
Section~\ref{sec:lag} explains this delay through the redundancy of
heads relative to the task; the pair's advantage in this regime is
speed of solution, not final accuracy, since all arms finish
between $.90$ and $.95$. As elsewhere, the result is specific to
the stated configuration and budget.

\emph{Third}, the length-generalisation and recognition results are a
diagnostic at small scale, not a statement about large models. The
fixed-precision theory says NoPE soft-attention transformers cannot
recognise even bounded Dyck languages, and places them exactly in the
past-only fragment of linear temporal logic, a proper subclass of the
regular languages \cite{li2025}; Del\'etang \emph{et al.}~\cite{deletang2023}
report the same boundary empirically; both findings were reproduced
here for both models. The out-of-distribution differences of
Section~\ref{sec:express} are underpowered at three seeds and would
need $n \ge 10$ seeds to resolve. The recognizer negatives were random strings rather
than minimally-unbalanced Dyck words. This may raise measured Dyck-2
accuracy and limits interpretation of the absolute scores.

\emph{Fourth}, ListOps here is a scaled-down diagnostic (depth $\le
4$, T$=$64), far from the Long Range Arena 2K-token, depth-10 setting
of \cite{tay2021}, where the original transformer baseline reaches
only $36.4\%$.
The $.60$--$.69$ scores are not LRA results; they compare the
architectures under the same training setup.

\emph{Fifth}, MQAR accuracy depends on training budget and seed
around the transition, so low fixed-budget accuracy does not by itself
show a representational limit. A preliminary single-seed pair-model
pilot at S $= 16$, $24$, $32$ remained near chance. The cyclic model
was not run in that pilot, so it does not establish a shared
long-sequence cliff; the final comparison is limited to
S $\in \{8, 10\}$.

\emph{Sixth}, the 7B extrapolation is an extrapolation. It has not
been measured, and this paper does not pretend otherwise.

\emph{Seventh}, the cyclic pairs variant (nearest cyclic pairs rather
than the wrap-around) is outside this experiment battery; no claim is
made about it.

\emph{Finally}, on the name: this paper's ``ring'' denotes the cyclic
head wiring of Fig.~\ref{fig:ring}. Ring Attention \cite{liu2024} is
a distributed context-parallel scheme and is unrelated.

\section{Conclusion}
\label{sec:concl}

The cyclic construction replaces the separate query and key
projections of each head with a cyclic arrangement of one projection
per head. This halves the query--key parameters while preserving
a family of head-specific bilinear scores. FastCycle adds a
parameter-free gate that closes part of the learning-speed gap.

The main weakness appears in highly redundant tasks, where the ring
can take longer to learn. In the measured phase-transition run,
the pair crosses about 540 steps earlier. Some wide MQAR settings
are difficult for both architectures, with the pair model also
failing to learn within the training budget. This distinction matters
in practice: multi-head transformers are generally used on richer
tasks rather than highly redundant retrieval problems. In that
regime, Section~\ref{sec:lag} predicts less or no delay.

On the real-world tasks we tested (AG News, CIFAR-10, and
character-level language modelling), the ring achieves performance
comparable to or better than the pair, with differences within
$0.1$ percentage points on AG News, $0.5$ on CIFAR-10, and $0.01$
in cross-entropy (Fig.~\ref{fig:realworld}). FastCycle stays in the
same band. These results show that the ring can retain competitive
performance while halving query--key parameters and reducing
projection computation. This also gives it the potential to run
faster than traditional multi-head attention, although actual speed
gains depend on the implementation and hardware.

The results are limited to small models and short contexts. Testing
the architecture at larger scale and on longer sequences remains
necessary.

\section*{Acknowledgement}

The reported measurements were run on a $2\times$RTX A4000 host.

\bibliographystyle{IEEEtran}
\bibliography{references}

\end{document}